\documentclass{article}

\usepackage{arxiv}
\usepackage{booktabs}
\usepackage[utf8]{inputenc} 
\usepackage[T1]{fontenc}    
\usepackage[]{hyperref}       
\usepackage{url}            
\usepackage{booktabs}       
\usepackage{amsfonts}       
\usepackage{nicefrac}       
\usepackage{microtype}      
\usepackage{tabularx}
\usepackage{lipsum}         
\usepackage{graphicx}
\usepackage{doi}
\usepackage{algorithm}
\usepackage[toc, acronym, automake,nonumberlist]{glossaries}
\setkeys{glslink}{hyper=false}
\usepackage{algpseudocode}
\usepackage{subfig}

\usepackage[noabbrev, capitalize]{cleveref}       
\makeglossaries
\newacronym{fox}{FOX}{FOX optimization algorithm}
\newacronym{fa}{FA}{Fox agent}
\newacronym{ga}{GA}{genetic algorithms}
\newacronym{pso}{PSO}{particle swarm optimization}
\newacronym{rl}{RL}{reinforcement learning}
\newacronym{mdp}{MDP}{markov decision processes}
\newacronym{ai}{AI}{artificial intelligence}
\newacronym{ba}{BA}{bees algorithm}
\newacronym{mse}{MSE}{mean squared error}
\newacronym{mae}{MAE}{mean absolute error}
\newacronym{ibl}{IBL}{instance-based learning}
\newacronym{caesar}{CAESAR}{Convergence-AwarE SAmpling with scReening}

\title{QF-tuner: Breaking Tradition in Reinforcement Learning}

\usepackage{authblk}

\newbox{\orcid}\sbox{\orcid}{\includegraphics[scale=0.06]{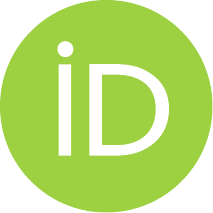}} 
\author[1]{%
	\href{https://orcid.org/0000-0002-6232-3900}{\usebox{\orcid}\hspace{1mm}Mahmood A. Jumaah\thanks{\texttt{cs.22.27@grad.uotechnology.edu.iq}}}%
}
\author[1]{%
	\href{https://orcid.org/0000-0002-7216-4149}{\usebox{\orcid}\hspace{1mm}Yossra H. Ali\thanks{\texttt{Yossra.H.Ali@uotechnology.edu.iq}}}%
}
\author[2]{%
	\href{https://orcid.org/0000-0002-8661-258X}{\usebox{\orcid}\hspace{1mm}Tarik A. Rashid\thanks{\texttt{tarik.ahmed@ukh.edu.krd}}}%
}
\affil[1]{Department of Computer Science, University of Technology, Baghdad 10066, Iraq}
\affil[2]{Department of Computer Science and Engineering; AIIC, University of Kurdistan Hewlêr, Erbil 44001, Iraq}

\renewcommand{\shorttitle}{\textit{QF-tuner}}

\begin{document}
\maketitle
\begin{abstract}
	In reinforcement learning algorithms, the hyperparameters tuning method refers to choosing the optimal parameters that may increase the overall performance. Manual or random hyperparameter tuning methods can lead to different results in the reinforcement learning algorithms. In this paper, we propose a new method called QF-tuner for automatic hyperparameter tuning in the Q-learning algorithm using the FOX optimization algorithm (FOX). Furthermore, a new objective function has been employed within FOX that prioritizes reward over learning error and time. QF-tuner starts by running the FOX and tries to minimize the fitness value derived from observations at each iteration by executing the Q-learning algorithm. The proposed method has been evaluated using two control tasks from the OpenAI Gym: CartPole and FrozenLake. The empirical results indicate that the QF-tuner outperforms other optimization algorithms, such as~\gls{pso},~\gls{ba},~\gls{ga}, and the random method. However, on the FrozenLake task, the QF-tuner increased rewards by 36\% and reduced learning time by 26\%, while on the CartPole task, it increased rewards by 57\% and reduced learning time by 20\%. Thus, the QF-tuner is an essential method for hyperparameter tuning in Q-learning algorithms, enabling more effective solutions to control task problems.
\end{abstract}

\section{Introduction} \label{sec: introduction}
\Gls{rl} is a branch of machine learning in which the agent learns how to make a perfect decisions by interacting with an environment. It aims to identify optimal actions based on the results of these interactions to develop strategies that maximize long-term rewards. Moreover,~\gls{rl} involves selecting the actions that achieve the maximum rewards in each state of the problem's environment. This behavior is learned through interactions with the environment by observing its trial-and-error responses. It is similar to how children explore their surroundings and learn from experiences to develop effective strategies through repeated practice~\cite{rl_strategy,rl_app3}.

A~\gls{mdp} is formally defined as a tuple $(S, A, P, R, \gamma)$. The $S$ represents a finite set of states, and $A$ denotes a finite set of actions available to an agent. When the agent selects an action $a \in A$ while in a current state $s \in S$, the probability of transitioning to the subsequent state $s' \in S$ is governed by the state-transition probability function $P : S \times A \times S \rightarrow [0, 1]$. Upon transitioning from state $s$ to state $s'$ after performing action $a$, the agent receives an immediate numerical reward defined by the reward function $R : S \times A \times S \rightarrow \mathbb{R}$. The discount factor $\gamma \in [0, 1)$ indicates how much the agent values immediate rewards in relation to future rewards. Thus, reflecting the importance assigned to long-term outcomes through the decision-making process~\cite{mdp}. Q-learning is considered a common algorithm in~\gls{rl}, which is model-free, off-policy, widely used, and simple to implement used to solve finite~\glspl{mdp}. It solves problems by transitioning between states based on an approximate action-value function. This approach allows the Q-learning to select actions that maximize cumulative rewards within the problem environment~\cite{qlearning}. However, Q-learning is utilized in many domains, including gaming, control systems, robotics, and task scheduling~\cite{rl_app1,rl_app2}.

The~\gls{fox} is a recent optimization algorithm proposed by Mohammed and Rashid in 2022.~\gls{fox} inspired by the behaviors of red foxes, such as walking, jumping, and hunting for prey. This algorithm mathematically simulates these natural behaviors. It works as a population-based, where multiple independent agents, known as~\glspl{fa}, explore potential solutions.~\gls{fox} has shown superior performance compared to established optimization algorithms like PSO, GA, GWO, FDO, and WOA~\cite{fox}. However, in~\gls{rl} algorithms, hyperparameters are set before the beginning of learning process. They used to control the learning behavior of the~\gls{rl} algorithm. They affect how the algorithm explores and exploit within the environment, such as the learning rate and discount factor~\cite{hp,foxann}. While~\gls{rl} algorithms can yield maximum rewards; they require overhead tuning the hyperparameter to achieve optimum performance~\cite{hp2}. The difficulty is in determining the most effective values for each hyperparameter, as even minor adjustments might influence the learning process in any~\gls{rl} algorithm. However, manual or random hyperparameter tuning can be complex and time-intensive process. Several methods have been suggested for hyperparameter tuning manually, randomly, or semi-automatic adjusting values for each parameter~\cite{hp_summary}. Furthermore, these methods may not generally apply to all~\gls{rl} problems due to environmental variables and conditions variations.

This paper employed the~\gls{fox} for automatic hyperparameter tuning in the Q-learning algorithm to find the optimal hyperparameters that results in maximum rewards and the lowest learning time (steps). This method, QF-tuner, solves the problems associated with manual hyperparameter tuning in~\gls{rl} algorithms. It offers a robust and effective solution that addresses the challenges of hyperparameter tuning. Furthermore, a new objective function has been proposed and utilized in the~\gls{fox}. it includes reward, error, and learning time and prioritizes reward over error and learning time. However, the contributions of this paper are as follows:
\begin{enumerate}
  \item QF-tuner uses the state-of-the-art FOX optimization algorithm for automatic hyperparameter tuning in Q-learning, addressing the complexities of manual or random tuning.
  \item A new objective function is proposed that prioritizes reward over error and learning time.
  \item Empirical results show that QF-tuner improves Q-learning performance, achieving higher rewards with fewer learning steps in control tasks like FrozenLake and CartPole.
\end{enumerate}

QF-tuner is evaluated using two control tasks from OpenAI Gym: FrozenLake and CartPole. FrozenLake presents a grid-world design where the agent must navigate slippery terrain to reach a goal while avoiding hazards, requiring decision-making in a stochastic environment. A CartPole balances a pole on a moving cart with precise control to maintain stability in a changing environment~\cite{gym}. These tasks are benchmarks for evaluating~\gls{rl} algorithms, offering many challenges that test the robustness of the proposed approach across different control situations. The empirical results show that QF-tuner has improved the overall Q-learning algorithm's performance by increasing the rewards with minimum steps.

This paper is structured as follows:~\cref{sec:related work} reviews the latest studies in hyperparameter tuning using optimization algorithms.~\cref{sec:materials and methods} discuss the OpenAI Gym control tasks and explain the Q-learning, \gls{fox}, and the QF-tuner in detail.~\cref{sec:results and discussion} list and discuss the experimental results. Finally,~\cref{sec:conclusion} presents the conclusion and future work suggestions.

\section{Related Work} \label{sec:related work}
In recent years, many investigations have focused on solving control tasks using~\gls{rl}, particularly with Q-learning and its variants. While Q-learning is a powerful algorithm for such tasks, its performance heavily relies on hyperparameter tuning, which can be a time-consuming and challenging process. Many existing approaches have attempted to reduce this burden, but they still face limitations in large-scale or complex control environments.

For instance, the authors of~\cite{r2} proposed a framework to reduce the burden of hyperparameter tuning by using random feature approximation for efficient state feature computation and a factorial policy for fast processing of discrete actions. Although their approach effectively reduced overhead for large-scale tasks, it does not directly address the challenge of automatically optimizing hyperparameters in Q-learning, which is still a bottleneck for performance improvement. Furthermore, the study by~\cite{related_cartpole1} suggested two Q-learning-based methods for the CartPole task. They used~\gls{mse} and~\gls{mae} as reward objective functions. While their results demonstrated that Q-learning with~\gls{mse} tends to ignore outliers, and~\gls{mae} gives more weight to outliers, these methods still require manual adjustment of hyperparameters for optimal performance, which can be inefficient to suboptimal solutions. Further advancements were made by the authors of~\cite{related_ga_deep_rl}, who proposed a scalable~\gls{ga} for hyperparameter exploration in deep~\gls{rl}. However, while this method was effective in maximizing rewards in fewer episodes, it still faced challenges in terms of computational cost and the ability to escape local optima in complex tasks. Additionally, some hybrid approaches, such as the combination of the~\gls{ba} and Q-learning in~\cite{related_Q_BA}, have been proposed to enhance feature selection in classification tasks, such as coronary heart disease diagnosis. However, while these hybrid models showed improved accuracy in specific domains, they still rely on fixed or semi-adaptive optimization strategies that struggle with the non-stationary nature of~\gls{rl} tasks.

In the case of the FrozenLake task,~\cite{related_frozenlake1} compared Q-learning to cognitive IBL algorithms and found that IBL outperformed Q-learning in terms of reduced learning time and better generalization. However, this study highlighted the challenge of developing more effective Q-learning-based methods that can handle such tasks without facing the same limitations in time efficiency and generalization. Moreover, the CAESAR aggregation scheme~\cite{related_frozenlake2} aimed to improve learning efficiency by blending convergence-aware sampling with a screening mechanism. Although it showed positive results in enhancing learning across different environments, it did not address the underlying issue of efficiently tuning the hyperparameters in Q-learning itself, which remains a key challenge for achieving optimal performance in control tasks.

Despite these advances, conventional hyperparameter tuning methods like~\gls{pso},~\gls{ga}, and~\gls{ba} often struggle to escape local optima, especially in the context of complex~\gls{rl} tasks~\cite{related_Q_BA, pso_Overview, related_Q_GA}. These methods rely on fixed or semi-adaptive strategies, which limit their effectiveness, particularly in dynamic or highly variable task environments. In contrast, the QF-tuner introduces a novel, automatic hyperparameter tuning approach for Q-learning that leverages the~\gls{fox}. This method overcomes the limitations of existing techniques by dynamically adjusting hyperparameters based on an objective function designed to prioritize reward maximization while minimizing learning error and time. By addressing the specific challenges of hyperparameter tuning in Q-learning, QF-tuner provides a more stable and efficient solution, particularly in tasks where other optimization algorithms struggle with convergence.

\section{Materials and Methods} \label{sec:materials and methods}
This section presents materials (control tasks) and the proposed method (QF-tuner). It starts by explaining the OpenAI Gym environment control tasks—FrozenLake and CartPole. Subsequent sections will provide a detailed background of the Q-learning algorithm,~\gls{fox}, and the QF-tuner. However, the notation list of the symbols used is provided in~\cref{tab:notation_list}.

\begin{table}[ht]
	\centering
	\caption{Notation List} \label{tab:notation_list}
	\begin{tabular}{ll}
		\hline
		\textbf{Symbol}   & \textbf{Description}                                           \\
		\hline
		$ \delta_t $      & Temporal difference error                                      \\
		$ R_{t+1} $       & Reward at time step $t+1$                                      \\
		$ \gamma $        & Discount factor                                                \\
		$ \alpha $        & Step size                                                      \\
		$ S_t $           & Current state                                                  \\
		$ S_{t+1} $       & Next state                                                     \\
		$ A_t $           & Current action                                                 \\
		$ a $             & Selected action                                                \\
		$ Q(S_t, A_t) $   & Action-value function for state $S_t$ and action $A_t$         \\
		$ Q^*(S_t, A_t) $ & Optimal action-value function for state $S_t$ and action $A_t$ \\
		$ \pi^* $         & Optimal policy                                                 \\
		$ T_i $           & Arbitrary number in the range [0, 1] for iteration $i$         \\
		$ S $             & Sound speed                                                    \\
		$ DSF_i $         & Sound distance from the~\gls{fa}                              \\
		$ DFP_i $         & Distance from prey                                             \\
		$ Jump_i $        & Jump height for the~\gls{fa}                                  \\
		$ X_{i+1} $       & Position of the~\gls{fa} at iteration $i+1$                   \\
		$ c_1 $           & Constant value for exploitation step 1                         \\
		$ c_2 $           & Constant value for exploitation step 2                         \\
		$ tt $            & Time average                                                   \\
		$ ca $            & Exploration control factor                                     \\
		$ \text{dim} $    & Dimension of the problem                                       \\
		$ st $            & Learning steps or time taken to reach the best solution        \\
		$ n $             & Number of runs for the tuning process                          \\
		$ g $             & Number of~\gls{fa}s                                           \\
		$ max\_iter $     & Maximum number of iterations for optimization                  \\ \hline
	\end{tabular}
\end{table}

\subsection{Materials} \label{sec:Materials}
This paper used two control tasks—FrozenLake and CartPole—developed by OpenAI Gym to evaluate the QF-tuner. In the FrozenLake control task, an agent has to cross a grid world over a FrozenLake. This task introduces obstacles, creating a challenging environment~\cite{frozenlake_for_Materials}. Furthermore, in the CartPole environment, an inverted pendulum on a moving cart needs to be stabilized. This task is common when evaluating the ability of~\gls{rl} algorithms to maintain stability in a continuous state space~\cite{cartpole_for_Materials}. However, QF-tuner uses these tasks to guarantee a standardized evaluation framework for evaluating the performance. Moreover, this paper advances experimental and makes relevant comparisons with previous research findings using other methods on these control task environments.

\subsection{Q-learning Algorithm}\label{sec:Q-learning Algorithm}
One of the most promising fields of~\gls{ai} is~\gls{rl}, which can learn directly from interactions between the agent and the environment~\cite{rl_and_ai} .\gls{rl} does not require learning data or a target label; it only needs to take action in the problem's environment by the agent and attempt to earn a reward~\cite{rl_strategy}. The agent's learning process mimics how humans play a specific video game. The agent always tries to win; if it loses, it will avoid the behavior that led to the loss in the next attempt~\cite{rl_in_games}. In 1989, Watkins introduced a method considered an early breakthrough in~\gls{rl} known as Q-learning, defined by the following equations~\cite{qlearning}:

\begin{equation}
	\delta_t = R_{t+1} + \gamma \max_a Q(S_{t+1}, a) - Q(S_t, A_t)
	\label{eq:temporal difference error}
\end{equation}

\begin{equation}
	Q(S_t, A_t) = Q(S_t, A_t) + \alpha \delta_t
	\label{eq:qlearning_update}
\end{equation}
Where $\delta$ is the temporal difference error, $R$ is the reward, $\alpha$ is the step size, $\gamma$ is the discount factor, $S_t$ is the current state, $S_{t+1}$ is the next state, $A_t$ is the current action, and $a$ is the selected action. The $\max_a$ ensures the use of the optimal Q-value of the next state. Moreover, $Q(S_t, A_t)$ is the action-value function representing the cumulative reward expected if action $A$ is taken in the state $S$. Action-value pairs are stored in a Q-table table with dimensions of [number of states, number of actions].

In the learning process, the Q-learning's agent obtains the reward $R$ after taking the action $A$ in the current state $S$, then moves to the next state $S_{t+1}$. Additionally, the Q-learning algorithm tries to find the best course of actions that follow policy in the~\gls{mdp} by determining the optimal function of action-value using the update rule from~\cref{eq:temporal difference error,eq:qlearning_update}~\cite{qlearning2,alaa2,qlearning3}. Moreover, the agent interacts with the environment to observe the reward and updates the action-value function to improve its policy estimates over time. Agent interactions process iteratively, allowing the algorithm to modify its policy based on observations of the problem's environment. Finally, the agent should have the ability to select the best action (optimal policy) in a given state using the estimated Q-table by the following~\cref{eq:optimal_policy}~\cite{optimalPolicy}:

\begin{equation}
	\pi^* = \arg\max_a Q^*(s, a)
	\label{eq:optimal_policy}
\end{equation}

where $\pi^*$ represents the optimal policy and $\arg\max_a$ is a function returns an action that maximizes the value of $Q^*(a) $ based on learned policy~\cite{optimalPolicy,uot_abeer}.

\subsection{FOX optimization algorithm}\label{sec:FOX optimization algorithm}
\gls{fox} is a new optimization algorithm inspired by red foxes' hunting behavior that aims to search for the
the best solution with the best fitness value~\cite{fox}. FOX has many search agents working together iteratively, each trying to find the best fitness value during the search. It involves two major phases: exploration and exploitation. In exploration, the~\gls{fa} uses a random walk strategy to locate and catch prey, utilizing its ability to detect ultrasound. During its search for prey, the~\gls{fa} may listen for sounds indicating where its prey is. In exploitation, the~\gls{fa} can hear ultrasound, so it takes time for the sound of its prey to reach it. The speed of sound is 343 meters per second~\cite{soundSpeed}. However, since this speed is always the same, FOX uses a different method to measure it. The~\gls{fa} jumps when it thinks it can catch its prey based on how long the sound takes to reach it. Thus, the~\gls{fa}'s ability to catch prey depends on the time it takes for the sound to reach them while jumping~\cite{related_Q_GWO}. The following equations described the primary steps of~\gls{fox} in detail.

\begin{equation}
	S = \frac{\text{Best}X}{T_i}
	\label{eq:soundSpeed}
\end{equation}

\begin{equation}
	DSF_i = S \times T_i
	\label{eq:dsf}
\end{equation}

The variable $ DSF_i$ represents the sound distance from the~\gls{fa}. The sound speed in the medium is represented by the variable $S$ and calculated by~\cref{eq:soundSpeed}. The variable $T$ represents an
arbitrary number in the $[0, 1]$ range. The variable $i$ represents the current iteration of the optimization process~\cite{optimizationYossra}. Moreover, The distance of a~\gls{fa} from its prey is defined as follows~\cite{used_fox_1}:

\begin{equation}
	DFP_i = DSF_i \times 0.5
	\label{eq:dfp}
\end{equation}

Once a~\gls{fa} has determined the distance between itself and its prey, it should jump and catch the prey accurately. The~\gls{fa} must calculate the required jump height using the~\cref{eq:jump} to perform this jump~\cite{fox, alaa1}.

\begin{equation}
	Jump_i = 0.5 \times 9.81 \times t^2
	\label{eq:jump}
\end{equation}
where 9.81 represents the acceleration due to gravity, and $t$ is the time average taken by sound to travel~\cite{fox}.

\begin{equation}
	X_{i+1} = DFP_i \times Jump_i \times c_1
	\label{eq:x_exploit_c1}
\end{equation}

\begin{equation}
	X_{i+1} = DFP_i \times Jump_i \times c_2
	\label{eq:x_exploit_c2}
\end{equation}
where the value of $c_1$ and $c_2$ are 0.180 and 0.820, respectively. These values have been determined based on the~\gls{fa}'s jump. The direction of the jump is randomly chosen to be either northeast or opposite direction. However, the~\gls{fa} continuously moves to a new position in exploration and exploitation. In exploitation, the new position is determined using either~\cref{eq:x_exploit_c1} or~\cref{eq:x_exploit_c2}.

On the other hand, the exploration determined the new position using the following equation~\cite{fox}:
\begin{equation}
	X_{i+1} = \text{Best}X \times \text{rnd}(1, \text{dim}) \times \min(tt) \times ca
	\label{eq:x_explore}
\end{equation}

The variable $tt$ is the time average, which is equal to the summation of the variable $T$ used in~\cref{eq:soundSpeed} divided by the dimension of the problem (dim). The exploration is controlled by variable $ ca $ that is computed by $2 \times [i-(i/max(iter))]$. However, the source code of~\gls{fox} implementation can be accessed by~\hyperlink{https://github.com/Hardi-Mohammed/FOX}{https://github.com/Hardi-Mohammed/FOX}. The next subsequent section will present an explanation of the QF-tuner.

\subsection{Hyperparameter Tuning Method (QF-tuner)}\label{sec:Hyperparameter Tuning Method (QF-tuner)}
The Q-learning algorithm has several hyperparameters that affect the agent's performance and behavior regarding the learned policy. From~\cref{eq:qlearning_update,eq:temporal difference error}, the following hyperparameters can be noticed: step size ($\alpha$) and discount factor ($\gamma$). They should be set before the beginning of the learning process. The $\alpha$ is crucial in updating the action-values function at each time step. It determines the extent to which new information overrides old information, and its value is typically within the range $[0, 1]$. A lower $\alpha$ value means that the algorithm will be less sensitive to new information and will take longer to converge, while a higher value results in faster learning but may lead to instability. Therefore, choosing the correct value of $\alpha$ is essential for achieving the best performance in any learning algorithm~\cite{hp_alpha}. In comparison, the $\gamma$ affects future reward for an agent's decisions and its value range $[0, 1]$. A more significant value (close to 1) indicates that the agent emphasizes future rewards, while a smaller value means that the agent gives less importance to future rewards. This parameter controls the balance between immediate and future rewards, affecting the agent's decision-making process~\cite{hp_gamma}.

Developing a multi-objective function was necessary to address the multiple objectives of increasing reward, reducing the learning error, and minimizing the learning time (steps). This function is needed to encapsulate the simultaneous optimization of these three objectives within a single fitness function. Thus, this paper suggested the following objective function:

\begin{equation}
	\textit{fitness} = \sum_{i=1}^{n} (2R - \delta) \times \frac{1}{st}
	\label{eq:fitness}
\end{equation}
where $\delta$ represents the temporal difference error computed by the~\cref{eq:temporal difference error} and the value of $st$ represents the amount of time (learning steps) the~\gls{fa} takes to reach the best solution. Indeed, this equation is calculated on the last quarter of episodes from the entire episodes to ensure the stability of the learning process, so $i = episodes - episodes/4$. For example, if the number of episodes  = 200, the fitness calculation starts from the episode 150.

\begin{figure}[ht]
	\centerline{\includegraphics[width=0.5\linewidth]{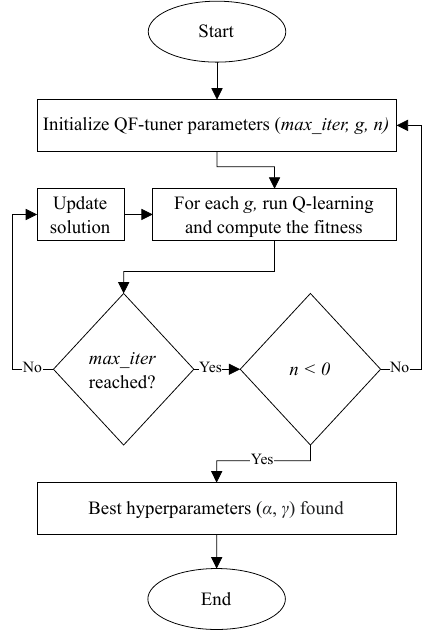}}
	\caption{Flowchart of the QF-tuner}
	\label{fig:QF-tuner}
\end{figure}

The QF-tuner employed the \gls{fox} with Q-learning algorithms in a new model that automates the tuning of hyperparameters without human intervention. Setting up the QF-tuner involves determining the number of iterations $max\_iter$, the number of~\gls{fa}s $g$, and the number of runs $n$, as illustrated in~\cref{fig:QF-tuner}. The~\gls{fox} is employed, and its initial solution serves as the initial hyperparameters ($\alpha$ and $\gamma$) of the Q-learning algorithm. Next, The Q-learning algorithm starts learning iteratively based on the number of episodes, during which reward, learning error, and convergence time values are computed to derive the fitness value using~\cref{eq:fitness}. The~\gls{fa}'s solution is updated if $max\_iter$  or appropriate convergence is not reached. Otherwise, the QF-tuner starts again until the number of runs ($n$) is achieved to prevent getting stuck in the local optima.
Finally, the optimal solution consists of the best set of hyperparameters ($\alpha$ and $\gamma$), providing robust and practical solutions for real-world problem-solving, without the need for manual or random tuning.

\section{Results and Discussion} \label{sec:results and discussion}
\begin{figure}[h]
	\centerline{\includegraphics[width=0.5\linewidth]{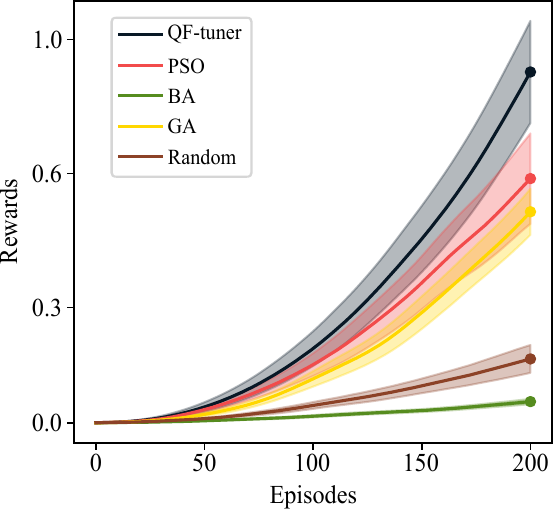}}
	\caption{Normalized rewards of the QF-tuner and other optimization algorithms on the FrozenLake task. The normalization allows for fair comparison of performance, highlighting the QF-tuner's superior reward within fewer learning steps.}
	\label{fig:frozenlake_resutls}
\end{figure}

\begin{figure}[h]
	\centerline{\includegraphics[width=0.5\linewidth]{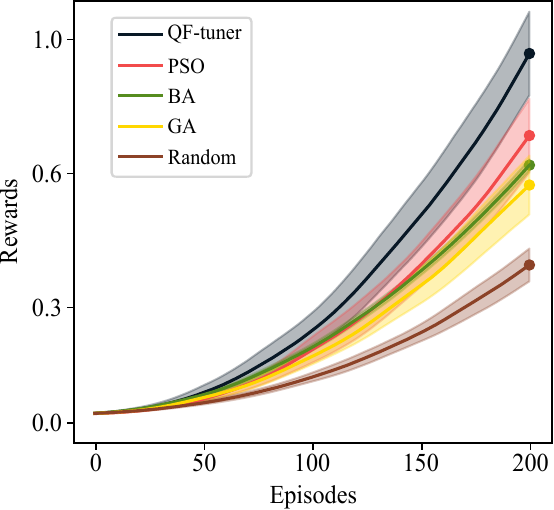}}
	\caption{Normalized rewards of the QF-tuner and other optimization algorithms on the CartPole task. Normalization facilitates comparison, showing QF-tuner's higher reward and reduced learning time.}
	\label{fig:cartpole_resutls}
\end{figure}

Experiments were conducted to evaluate the performance of the QF-tuner and compare it with several established optimization algorithms, including \gls{pso}, \gls{ba}, and \gls{ga}. OpenAI Gym environment control tasks were utilized during the experiments, as explained in~\cref{sec:Materials}. The parameter settings for the QF-tuner comprised 100 iterations, 30~\gls{fa}s, 10 runs, and 200 episodes, with 200 steps for the FrozenLake control task and 500 steps for the CartPole control task. The following~\cref{tab:frozenlake_results,tab:cartpole_results,fig:frozenlake_resutls,fig:cartpole_resutls} display the performance results using hyperparameters (Alpha $\alpha$, and Gamma $\gamma$), as well as Rewards, where the maximum value for FrozenLake is one and for CartPole is 500. The processing time (Time) was also considered to measure the speed of the optimization algorithm. The reward curves in the~\cref{fig:frozenlake_resutls,fig:cartpole_resutls} have been normalized between 0 and 1 to reduce the variance and improve visualization.

In~\cref{fig:frozenlake_resutls}, the QF-tuner achieved a reward of 0.95, surpassing~\gls{pso} (0.78),~\gls{ba} (0.65),~\gls{ga} (0.62), and Random (0.34).~\cref{tab:frozenlake_results} shows that QF-tuner reduced learning time (steps) by 33\% compared to~\gls{ga}, 25\% compared to~\gls{pso}, and 20\% compared to~\gls{ba}. Furthermore, the average increase in the reward for QF-tuner was about 36\% compared to the other optimization algorithms, while the average reduction in learning time was around 26\%, demonstrating its superior performance in the FrozenLake control task.
\begin{table}[h]
	\centering
	\caption{Empirical results using the QF-tuner and other optimization algorithms on FrozenLake control task.}
	\begin{tabularx}{\linewidth}{XXXXX}
		\hline
		\textbf{Method}   & \textbf{Alpha} & \textbf{Gamma} & \textbf{Reward} & \textbf{Time (s)} \\
		\hline
		\textbf{QF-tuner} & \textbf{0.74}  & \textbf{0.97}  & \textbf{0.95}   & \textbf{156}      \\
		PSO               & 1.00           & 0.38           & 0.78            & 210               \\
		BA                & 0.94           & 0.64           & 0.65            & 197               \\
		GA                & 0.34           & 0.83           & 0.62            & 234               \\
		Random            & 0.09           & 0.74           & 0.34            & -                 \\
		\hline
	\end{tabularx}
	\label{tab:frozenlake_results}
\end{table}

In~\cref{fig:cartpole_resutls}, the QF-tuner achieved a reward of 499.08, surpassing~\gls{pso} (398.99),~\gls{ga} (215.03), Random (129.46), and~\gls{ba} (96.65).~\cref{tab:cartpole_results} shows that QF-tuner reduced learning time (steps) by 26\% compared to~\gls{ga}, 20\% compared to~\gls{pso}, and 14\% compared to~\gls{ba}. Moreover, the average gain in the reward for QF-tuner was approximately 57\% compared to the other optimization algorithms, while the average decrease in learning time was about 20\%, showing its superior performance in the CartPole control task.
\begin{table}[h]
	\centering
	\caption{Empirical results using the QF-tuner and other optimization algorithms on CartPole control task.}
	\begin{tabularx}{\linewidth}{XXXXX}
		\hline
		\textbf{Method}   & \textbf{Alpha} & \textbf{Gamma} & \textbf{Reward} & \textbf{Time (s)} \\
		\hline
		\textbf{QF-tuner} & \textbf{0.83}  & \textbf{0.95}  & \textbf{499.08} & \textbf{314}      \\
		PSO               & 1.00           & 0.67           & 398.99          & 394               \\
		GA                & 0.84           & 0.71           & 215.03          & 428               \\
		Random            & 0.56           & 0.38           & 129.46          & -                 \\
		BA                & 0.99           & 0.99           & 96.65           & 367               \\
		\hline
	\end{tabularx}
	\label{tab:cartpole_results}
\end{table}

Additionally, QF-tuner outperformed the results of~\cite{related_frozenlake1} by 8\% in reward on the $4 \times 4$ FrozenLake control task. Moreover, QF-tuner achieved a 6\% higher reward and a 26\% reduction in learning time compared to the study in~\cite{related_frozenlake2} on the same task. On the other hand, QF-tuner outperformed the results of~\cite{related_cartpole1} by 25\% in reward on the CartPole control task. Thus, these results indicate that the QF-tuner improves Q-learning performance in the FrozenLake and CartPole control tasks. The improvement in reward highlights the critical role of the proposed objective function in~\cref{eq:fitness} in boosting the optimization process. Furthermore, the significant reduction in learning time emphasizes the effective integration of FOX with Q-learning, leading to superior performance due to their strong compatibility. However, a key limitation of the QF-tuner is its resource-intensive nature, which prevents it from being directly applied to real-world problems. Instead, it needs to be executed in a simulated environment to fine-tune the hyperparameters, which can then be utilized directly.

\section{Conclusion}\label{sec:conclusion}
In conclusion, this paper presents the QF-tuner, a novel method for automatic hyperparameter tuning in Q-learning using the FOX optimization algorithm. A key contribution is the development of a new objective function that combines reward, learning error, and learning time with an emphasis on maximizing reward.  The QF-tuner was evaluated on the OpenAI Gym control tasks, FrozenLake and CartPole. The results show that the QF-tuner outperformed established algorithms, achieving a reward of 0.95 on FrozenLake, surpassing PSO (0.78), GA (0.62), and BA (0.65), while reducing learning time by 26\%. On CartPole, it achieved 499.08 reward, surpassing PSO (398.99) and GA (215.03), and reducing learning time by 20\%. However, tuning hyperparameters is time-intensive and requires multiple optimization iterations to reach the best set, which restricts its use in simulation environments. Future research can improve the proposed method, making it suitable for real-world applications and investigating alternative optimizers used in the Q-learning algorithm.

\bibliographystyle{elsarticle-num}
\bibliography{references}
\end{document}